\pdfoutput=1

\documentclass[11pt]{article}

\usepackage{acl}

\usepackage{times}
\usepackage{latexsym}

\usepackage[T1]{fontenc}

\usepackage[utf8]{inputenc}

\usepackage{microtype}

\usepackage{amsmath,amsfonts,bm}
\usepackage{hyperref}
\usepackage{url}
\usepackage{booktabs}
\usepackage{graphicx}
\usepackage{multirow}

\newcommand{\hs}[1]{\hspace{#1\tabcolsep}}

\title{Simple and Effective Unsupervised Speech Translation}

\author{
Changhan Wang, Hirofumi Inaguma, Peng-Jen Chen, Ilia Kulikov, Yun Tang \\\textbf{Wei-Ning Hsu, Michael Auli, Juan Pino}\\
Meta - Fundamental AI Research (FAIR) \\
\texttt{\{changhan,hirofumii,pipibjc,kulikov,yuntang,}\\\texttt{wnhsu,michaelauli,juancarabina\}@meta.com}
}

\begin{document}
\maketitle
\begin{abstract}
The amount of labeled data to train models for speech tasks is limited for most languages, however, the data scarcity is exacerbated for speech translation which requires labeled data covering two different languages.
To address this issue, we study a simple and effective approach to build speech translation systems without labeled data by leveraging recent advances in unsupervised speech recognition, machine translation and speech synthesis, either in a pipeline approach, or to generate pseudo-labels for training end-to-end speech translation models.
Furthermore, we present an unsupervised domain adaptation technique for pre-trained speech models which improves the performance of downstream unsupervised speech recognition, especially for low-resource settings. 
Experiments show that unsupervised speech-to-text translation outperforms the previous unsupervised state of the art by 3.2 BLEU on the Libri-Trans benchmark, on CoVoST 2, our best systems outperform the best supervised end-to-end models (without pre-training) from only two years ago by an average of 5.0 BLEU over five X-En directions.
We also report competitive results on MuST-C and CVSS benchmarks.
\end{abstract}

\section{Introduction}

Training supervised speech systems requires large amounts of labeled data which is often not available for all but a small fraction of the over 7,000 languages spoken around the world~\citep{lewis2022ethnologue}.
Despite much recent effort in creating speech translation corpora~\citep{di2019must,wang2021covost}, only a few dozen language directions are covered.
The lack of labeled training data is even more acute for speech translation because it requires aligned labeled data in two languages which increases the effort to create such datasets.
This poses the question of whether speech translation systems can be built using less labeled data or no labeled data at all.

Recent work on unsupervised speech recognition has achieved performance that can enable useful systems using no labeled data~\citep{yeh2018unsupervised,liu2018completely,chen2019completely,baevski2021unsupervised,liu2022towards}, enabled in large part by the advances in self-supervised speech representation learning~\citep{schneider2019wav2vec,baevski2020wav}.
These techniques were also used to build unsupervised text-to-speech systems~\citep{liu2022simple}.
Similarly, unsupervised text-to-text machine translation has shown great promise for certain language directions~\citep{conneau2018unsupmt,lample2018unsupervised,artetxe2018unsupervised}.

In this paper, we study a method to build end-to-end unsupervised speech-to-text and speech-to-speech translation systems trained on synthetic training data obtained by cascading existing unsupervised techniques:
we first transcribe speech utterances in the source language using unsupervised speech recognition~\citep{baevski2021unsupervised,liu2022towards}, then translate the resulting transcription using unsupervised machine translation~\citep{lample2018unsupervised,artetxe2018unsupervised,liu2020multilingual}, and finally synthesize the translation into a target language speech utterance using unsupervised speech synthesis~\citep{liu2022simple}.
We also consider applying the pipeline directly at inference time.
Our approach benefits from the use of self-supervised speech models~\citep{baevski2020wav,liu2020multilingual} and to further improve performance, we present a technique to adapt existing self-supervised models to the target domain. 



\begin{figure*}[t]
    \centering
    \includegraphics[width=1.9\columnwidth]{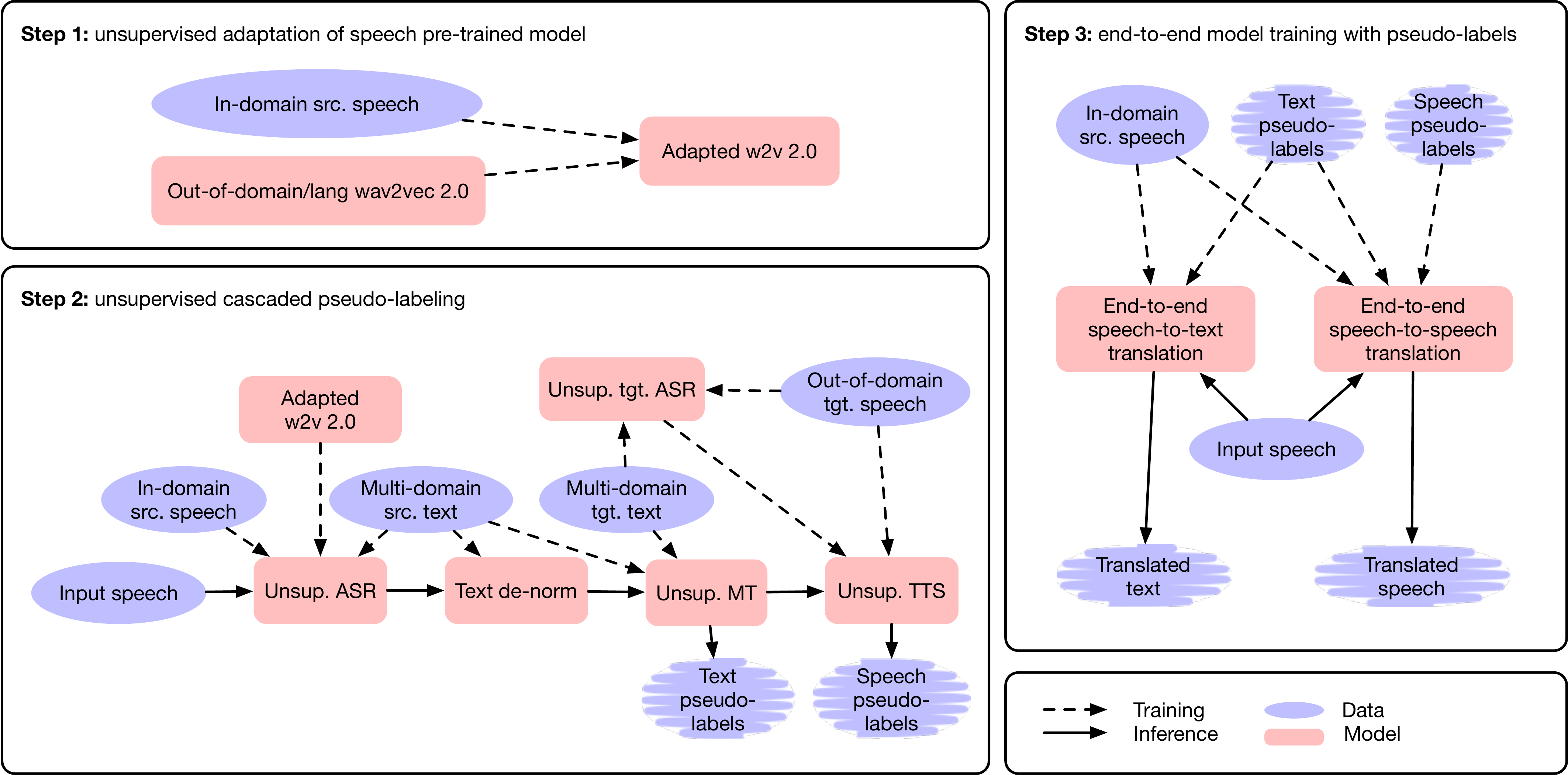}
    \caption{Overview of the proposed approach to unsupervised speech-to-text translation (S2TT) and speech-to-speech translation (S2ST). We first adapt speech pre-trained model (wav2vec 2.0) for the input language and domain of interest, and then cascade unsupervised speech recognition (ASR), unsupervised text de-normalization, unsupervised machine translation (MT) and unsupervised speech synthesis (TTS) models to produce pseudo-labels for end-to-end S2TT and S2ST model training. Our models rely only on unlabeled speech data and unpaired text data without the need of any human annotation.
    }
    \label{fig:overview}
\end{figure*}

\section{Background}
\paragraph{Unsupervised speech recognition.}
\citet{liu2018completely} presents some of the earliest work on unsupervised phoneme recognition and their work applies adversarial training. Wav2vec-U~\citep{baevski2021unsupervised} effectively applied self-supervised speech representations, introduced a new evaluation metric and compared to state-of-the-art supervised systems trained on large amounts of labeled data. Wav2vec-U 2.0~\citep{liu2022towards} simplifies audio-side pre-processing and improves accuracy through better architecture as well as better training objective.
\citet{lin2022analyzing} shows that out-of-domain speech pre-training or out-of-domain text data hurts the training robustness of Wav2vec-U models, especially under low-resource settings. 

\paragraph{Unsupervised speech synthesis.}
Recent work has demonstrated unsupervised speech synthesis systems to be able to achieve comparable performance to supervised systems~\citep{liu2022simple,ni2022unsupervised}.
The systems are trained on data resulting from labeling speech audio data with unsupervised speech recognition models and training text-to-speech models on the resulting models.

\paragraph{Unsupervised machine translation.}
\citet{lample2018unsupervised} and~\citet{artetxe2018unsupervised} built the first fully unsupervised machine translation (MT) systems by exploiting cross-lingual similarity of representations in multilingual sequence-to-sequence models, as well as back-translation for further refinements of the initial models. mBART~\citep{liu2020multilingual} used a similar model architecture and training process to build unsupervised MT models, but it utilized a larger-scale multilingual text corpus~\citep{conneau-etal-2020-unsupervised} and an updated noising strategy for pre-training with denoising autoencoder objective.


\paragraph{End-to-end speech translation.}
End-to-end sequence-to-sequence modeling has witnessed increased applications in speech-to-text translation~\citep{duong2016attentional,berard2016listen,weiss2017sequence,bansal2017towards,vila2018end,di2019adapting,ren2020simulspeech,li-etal-2021-multilingual} and speech-to-speech translation~\citep{Jia2019DirectST,kano2021transformer,pmlr-v162-jia22b}. Compared to cascaded systems, end-to-end speech translation models have simpler pipeline and lower inference latency. It is shown that recent end-to-end speech-to-text translation (S2TT) models perform comparably to the cascaded counterparts on the well-established MuST-C benchmark~\citep{bentivogli2021cascade}. Given the scarcity of speech translation corpora, there are recent attempts on building end-to-end S2TT models under low-resource settings~\citep{bansal2018low,bansal2019pre,cheng21_interspeech} or unsupervised settings~\citep{chung2019towards}.

\section{Methods}

Figure~\ref{fig:overview} provides an overview of our proposed approach to unsupervised speech-to-text translation (S2TT) and speech-to-speech translation (S2ST). 
We leverage a cascade of unsupervised models to produce pseudo-labels for end-to-end S2TT and S2ST model training. 
To mitigate language and domain mismatch in speech pre-training (wav2vec 2.0), we finetune wav2vec 2.0 models using unlabeled in-domain speech data, and then use the adapted models to build downstream speech recognition models.

\subsection{Unsupervised Cascaded Pseudo-Labeling}
\label{sec:cascade}
We cascade unsupervised speech recognition (ASR), 
unsupervised text de-normalization (TDN) and unsupervised machine translation (MT) models to produce pseudo-labels for S2TT. 
For S2ST, we additionally apply unsupervised speech synthesis (TTS) models to MT model outputs to obtain synthesized target speech.

\paragraph{Unsupervised ASR.} We adopt wav2vec-U 2.0~\citep{liu2022towards}, which learns a mapping from self-supervised speech representations to phonemes via adversarial training and decodes phonemes into words via a weighted finite state transducer~\citep{mohri1997finite}. To improve adversarial training stability and suppress overfitting in the low-resource settings, we add Gaussian noise to the frozen input features $X$
\begin{equation*}
X^\prime = X + {\cal N}(0, \sigma^2)
\end{equation*}
as well as R-Drop regularization~\citep{wu2021r} to the logit outputs of the generator
\begin{align*}
    {\cal L}_{rdp} &= \frac{1}{2}{\cal D}_{KL} ({\cal G}_1(X^\prime)~||~{\cal G}_2(X^\prime)) \\
    & + \frac{1}{2}{\cal D}_{KL} ({\cal G}_2(X^\prime)~||~{\cal G}_1(X^\prime))
\end{align*}
where ${\cal G}_1$ and ${\cal G}_2$ are two generator instances with different dropout masks, and ${\cal D}_{KL}$ is the Kullback-Leibler (KL) divergence. 
We add weighted $\alpha{\cal L}_{rdp}$ to the wav2vec-U 2.0 objective function, where $\alpha$ is a hyper-parameter. 
After adversarial learning, we follow~\citet{baevski2021unsupervised} to perform self-training with a Hidden Markov Model (HMM), and fine-tune the adapted wav2vec 2.0 model again with the CTC objective on the HMM labels. We denote the final ASR model as ``w2vu2-CTC''.

\paragraph{Unsupervised MT.} 
We adopt mBART~\citep{liu2020multilingual}, which has a Transformer architecture~\citep{vaswani2017attention} with model parameters shared across all training languages. 
It first obtains initial cross-lingual alignments for all languages via a denoising autoencoder objective~\citep{vincent2010stacked}, and then refines the alignments for one specific language pair via bidirectional online back-translation on that pair of languages. We denote this model as ``mBART-OBT''.

\paragraph{Unsupervised TDN.} 
\label{sec:unsup_adap}
ASR models decode normalized spoken-form texts, which have no case or punctuation (except hyphen and apostrophe). MT models, however, encode unnormalized written-form texts that have case and punctuation. This discrepancy leads to quality degradation when we cascade the two models directly for pseudo-labeling. To mitigate the mismatch, we de-normalize ASR model outputs into their unnormalized written form before feeding them into MT models. The text de-normalizer is a mBART model pre-trained with denoising autoencoder objective and fine-tuned with paired data of raw text (output) and its normalized version (input). 

\paragraph{Unsupervised TTS.} We follow~\citet{liu2022simple} to produce phoneme labels for unlabeled speech data with wav2vec-U 2.0, and then train an autoregressive Transformer TTS model~\citep{Li_Liu_Liu_Zhao_Liu_2019} on the pseudo-labeled data. For wav2vec-U 2.0, we perform HMM-based self-training and fine-tune pre-trained wav2vec 2.0 model with HMM phoneme labels.
To alleviate under-generation and over-generation issues in autoregressive models, we add R-Drop style consistency loss 
$${\cal L}_{c} = ||{\cal P}^{EOS}_1(X) - {\cal P}^{EOS}_2(X)||_1$$
to the objective function (weighted by a hyperparameter $\alpha$) for better end-of-sentence (EOS) predictions, where ${\cal P}^{EOS}_1$ and ${\cal P}^{EOS}_2$ are two EOS predictions on the same input $X$ with different dropout masks.



\subsection{Unsupervised Adaptation of wav2vec 2.0 Pre-trained Models}

Next, we present a method to improve performance when the domain of the data used for self-supervised pre-training differs from the downstream task domain which is often the case for low-resource languages.
Specifically, we adapt out-of-domain or out-of-language wav2vec 2.0 models to the domain and language of interest by fine-tuning the entire wav2vec 2.0 models on discrete labels obtained from unlabeled in-domain data using the CTC objective~\citep{graves2006connectionist}.

To obtain discrete labels, we first collect all the wav2vec 2.0 speech representations for the training data, and perform k-means clustering to identify $K$ clusters. 
Then for each utterance, we label each of its $T$ speech representation frames $\textbf{x}_t$ by the corresponding cluster ids $y_t\in\{1,...,K\}$, where $t\in \{1, ..., T\}$. 
Finally, we merge identical consecutive $y_t$ to obtain the final labels $y^\prime_{t^\prime}$, where $t^\prime\in\{1, ..., T^\prime\}$ and $T^\prime\le T$. 

After unsupervised fine-tuning with discrete labels, we discard the output projection layer used for the CTC objective, and use the resulting wav2vec 2.0 trunk instead of the original wav2vec 2.0 model in the downstream tasks. The adapted models are used to extract speech representations for wav2vec-U 2.0 models, as well as pre-train encoders of the CTC models in wav2vec-U self-training.

\subsection{End-to-end Model Training with Pseudo-labels}
After obtaining pseudo-labels from the cascade of unsupervised models, we train end-to-end S2TT and S2TT models with supervised objectives on these pseudo-labels. For end-to-end S2TT, we adopt the model architecture in~\citet{li-etal-2021-multilingual}, which we denote as ``w2v2-mBART''. We pre-train its encoder by the unsupervised ASR model, w2vu2-CTC, and pre-train its decoder by the unsupervised MT model, mBART-OBT. For end-to-end S2ST, we adopt a variant of Translatotron 2~\citep{pmlr-v162-jia22b}, Spec-T2,
which adds an additional encoder in between Translatotron 2's two decoders, and replace Translatotron 2's second decoder by an autoregressive Transformer decoder~\citep{Li_Liu_Liu_Zhao_Liu_2019}. Similar to w2v2-mBART, we pre-train Spec-T2's first encoder and first decoder by w2vu2-CTC and mBART-OBT, respectively. 

\begin{table*}[t]
    \small
    \centering
    \begin{tabular}{lcccccc}
    \toprule
     & Fr-En & Es-En & Ru-En & Et-En & Lv-En & \multirow{2}{*}{Avg.} \\
    Duration (hrs) & 264 & 113 & 16 & 3 & 2 & \\
    \midrule
    \midrule
    \multicolumn{7}{c}{\textit{Bilingual setup}} \\
    \textbf{Supervised learning + pre-training} \\
    \hspace{1.5mm}End-to-end (w2v2-mBART) & 35.7 & 36.2 & 39.4 & \phantom{a}5.7 & 13.5 & 26.1 \\
    \textbf{Supervised learning} \\
    \hspace{1.5mm}End-to-end (S2T Transformer;~\citealt{wang-etal-2020-fairseq}) & 26.3 & 23.0 & 14.8 & \phantom{a}0.1 & \phantom{a}2.5 & 13.3 \\
    \textbf{Unsupervised learning} \\
    \hspace{1.5mm}Cascaded (ASR$\rightarrow$TDN$\rightarrow$MT) & 24.4 & 23.4 & 27.8 & \phantom{a}8.5 & \phantom{a}7.6 & 18.3 \\
    \hspace{1.5mm}End-to-end (w2v2-mBART) & 24.2 & 24.0 & 25.6 & \phantom{a}3.9 & \phantom{a}2.8 & 16.1 \\
    \midrule
    \multicolumn{7}{c}{\textit{Multilingual setup}} \\
    \textbf{Supervised learning + pre-training} \\
    \hspace{1.5mm}End-to-end (w2v2-mBART), 21 langs.$\rightarrow$En~\citep{babu2021xls} & 32.9 & 34.1 & 26.4 & \phantom{a}3.5 & \phantom{a}6.0 & 20.6 \\
    \textbf{Supervised learning} \\
    \hspace{1.5mm}End-to-end (S2T Transformer), 21 langs.$\rightarrow$En~\citep{wang-etal-2020-fairseq} & 26.9 & 26.3 & \phantom{a}9.6 & \phantom{a}0.4 & \phantom{a}0.6 & 12.8 \\
    \textbf{Unsupervised learning} \\
    \hspace{1.5mm}End-to-end (w2v2-mBART), $\{\text{Fr,Es,Ru,Et,Lv}\}\rightarrow\text{En}$ & 24.3 & 24.0 & 22.8 & \phantom{a}3.1 & \phantom{a}1.0 & 15.0 \\
    \bottomrule
    \end{tabular}
    \caption{Bilingual and multilingual X-En \textbf{speech-to-text translation} results: test BLEU on CoVoST 2. 
    Et-En and Lv-En are low-resource with only 3h and 2h of training data, respectively. End-to-end modeling on these two directions suffers from overfitting.
    }
    \label{table:s2t_x_en}
\end{table*}

\begin{table}[t]
    \small
    \centering
    \begin{tabular}{l@{\hs{1.3}}c@{\hs{1.3}}c@{\hs{1.3}}c}
    \toprule
     & En-Es & En-Ru & En-Fr \\
     Duration (hrs) & 504 & 489 & 100 \\
    \midrule
    \midrule
    \multicolumn{4}{l}{\textbf{Supervised learning + pre-training}} \\
    \hspace{1.5mm}End-to-end (w2v2-mBART) & 32.4 & 20.0 & 23.1 \\
    \midrule
    \textbf{Supervised learning} \\
    \hspace{1.5mm}End-to-end (S2T Transformer) & 27.2$^\dagger$ & 15.3$^\dagger$ & 11.4 \\
    \midrule
    \textbf{Unsupervised learning} \\
    \hspace{1.5mm}\citet{chung2019towards}$^\ddagger$ & 
    N/A & N/A & 12.2 \\
    \hspace{1.5mm}Cascaded (ASR$\rightarrow$TDN$\rightarrow$MT) & 22.0 & 10.0 & 15.4 \\
    \hspace{1.5mm}End-to-end (w2v2-mBART) & 23.8 & 9.8 & 15.3 \\
    \bottomrule
    \end{tabular}
    \caption{Bilingual En-X \textbf{speech-to-text translation} results: test BLEU on MuST-C (En-Es and En-Ru) and Libri-Trans (En-Fr). Our best system outperforms previous state of the art~\citep{chung2019towards} on Libri-Trans by 3.7 BLEU. $^\dagger$~\citet{wang-etal-2020-fairseq}. $^\ddagger$ We report the $\text{S}_\text{libri}\mbox{-}\text{T}_\text{libri}$ + $\text{LM}_\text{wiki}$ + $\text{DAE}_\text{wiki}$ configuration with the best result selected supervisedly out of 10 runs.
    }
    
    \label{table:s2t_en_x}
\end{table}

\section{Experimental Setup}

We evaluate our translation models on 5 directions into English (Fr-En, Es-En, Ru-En, Et-En and Lv-En) and 3 directions out of English (En-Es, En-Ru and En-Fr). The 5 non-English languages are from 4 different Indo-European language family sub-groups: Romance (Fr and Es), Slavic (Ru), Uralic (Et) and Baltic (Lv). For the X-En directions, we evaluate S2TT models on CoVoST 2~\citep{wang2021covost} and evaluate S2ST models on CVSS-C~\citep{jia2022cvss}, which adds synthetic target speech to CoVoST 2 with a single canonical speaker voice. For the En-X directions, we only evaluate S2TT models. We use MuST-C~\citep{di2019must} for En-Es and En-Ru, as well as Libri-Trans~\citep{kocabiyikoglu2018augmenting} for En-Fr. For Libri-Trans, we follow~\citet{chung2019towards} to combine validation set and test set for evaluation.

\paragraph{Speech pre-training.} We use robust wav2vec 2.0~\citep{hsu21_interspeech} for English speech, which is trained on datasets from multiple domains. For non-English speech, we adapt open-source VoxPopuli\footnote{https://github.com/facebookresearch/voxpopuli}~\citep{wang2021voxpopuli} models by CTC fine-tuning with 1024 discrete labels (Fr, Es and Ru) or 128 discrete labels (Et and Lv). We use monolingual VoxPopuli models for Fr and Es, and multilingual models of similar languages for Ru, Et and Lv (Slavic, Uralic and Baltic languages, respectively). We extract speech representations from the 15-th layer of the original wav2vec 2.0 models for computing discrete labels.

\paragraph{Speech recognition.} For wav2vec-U 2.0 models, we extract speech representations from the 19-th (15-th) layer of the adapted (original) wav2vec 2.0 models. We increase the dropout on the batch normalized input features to 0.2. We set $\sigma=0.1$ for input Gaussian noise and $\alpha=1.0$ for R-Drop regularization. For wav2vec-U 2.0 loss weights, we set $\eta=3$ and choose $\lambda$, $\gamma$ and $\delta$ from 1.0 / 1.5, 1.5 / 2.5 and 0.3 / 0.5, respectively. For text data, we use open web crawled corpus, CC-100~\citep{conneau-etal-2020-unsupervised}, which is created with little curation and has large language coverage.
For supervised baselines, we fine-tune adapted wav2vec 2.0 models with CTC objective on labeled data, which we denote as ``w2v2-CTC''.

\paragraph{Machine translation.} We use CC-100~\citep{conneau-etal-2020-unsupervised} to train bilingual mBART \textit{large} models for each language pair. For bidirectional online back-translation, we use the same CC100 data and follow~\citet{liu2020multilingual} to apply 99\% vocabulary masking for the first 500 updates. For supervised baselines, we fine-tune mBART models with labeled data, which we denote as ``mBART-FT''.

\paragraph{Speech synthesis.} We train Transformer models (with ${\cal L}_c$ weight $\alpha=1.0$) on CVSS-C target speech from the It-En direction to avoid content overlaps with the selected 5 directions. For grapheme-to-phoneme conversion, we employ g2pE~\citep{g2pE2019} for English texts and Phonemizer~\citep{phonemizer2015} with espeak-ng\footnote{https://github.com/espeak-ng/espeak-ng} backend for texts in other languages. We resample audios to 22,050Hz and extract log-Mel spectrogram with FFT size 1024, window length 1024 and hop length 256.

\paragraph{End-to-end speech translation.} For bilingual S2TT, we pre-train its encoder/decoder with w2vu2-CTC/mBART-OBT for unsupervised models, or with w2v2-CTC/mBART-FT for supervised models that leverage pre-training. To alleviate overfitting in low-resource settings (Ru-En, Et-En and Lv-En), we duplicate training examples and equip them with 2 different pseudo-labels from mBART-OBT beam search decoding. For multilingual S2TT and S2ST, we pre-train speech encoder with XLS-R 0.3B~\citep{babu2021xls}, and pre-train text decoder with mBART-OBT from the En-Fr direction.

\paragraph{Checkpoint selection and averaging.} For unsupervised ASR, we adopt the unsupervised metric in~\citet{baevski2021unsupervised} and average the best 2 checkpoints in the same run. For unsupervised MT and unsupervised TTS, we average the last 5 checkpoints. For end-to-end S2TT/S2ST, we sort checkpoints by losses on the pseudo-labeled validation set and average the best 5 checkpoints.

\paragraph{Automatic evaluation of speech outputs.} Following a common practice, we first transcribe English speech outputs from the TTS or S2ST model with an open-source English ASR model\footnote{https://github.com/facebookresearch/fairseq/tree/main/\newline
examples/wav2vec (``Wav2Vec 2.0 Large (LV-60) + Self Training'')}, and then calculate WER or BLEU on the ASR transcription for automatic evaluation scores.

\begin{table*}[t]
    \small
    \centering
    \begin{tabular}{lcccccc}
    \toprule
    & Fr-En & Es-En & Ru-En & Et-En & Lv-En & \multirow{2}{*}{Avg.} \\
    Source duration (hrs) & 264 & 113 & 16 & 3 & 2 \\
    \midrule
    \midrule
    \textbf{Supervised learning + pre-training} \\
    \hspace{1.5mm}End-to-end (Spec-T2),
    $\{\text{Fr,Es,Ru,Et,Lv}\}\rightarrow\text{En}$ & 31.8 & 32.3 & 32.9 & \phantom{a}5.2 & \phantom{a}7.5 & 21.9 \\
    \midrule
    \textbf{Supervised learning} \\
    \hspace{1.5mm}End-to-end (Spec-T2), $\{\text{Fr,Es,Ru,Et,Lv}\}\rightarrow\text{En}$ & 27.4 & 27.7 & 25.4 & \phantom{a}4.1 & \phantom{a}2.5 & 17.4 \\ 
    \midrule
    \textbf{Unsupervised learning} \\
    \hspace{1.5mm}Cascaded (ASR$\rightarrow$TDN$\rightarrow$MT$\rightarrow$TTS), bilingual & 21.6 & 21.2 & 25.3 & \phantom{a}7.2 & \phantom{a}7.7 & 16.6 \\
    \hspace{1.5mm}End-to-end (Spec-T2), $\{\text{Fr,Es,Ru,Et,Lv}\}\rightarrow\text{En}$ & 21.2 & 20.1 & 19.9 & \phantom{a}3.2 & \phantom{a}2.8 & 13.4 \\
    \bottomrule
    \end{tabular}
    \caption{Multilingual X-En \textbf{speech-to-speech translation} results: test BLEU on CVSS-C. Our multilingual model is trained on a subset of 5 directions out of the 21 available directions. 
    Appendix~\ref{app:cvss_baselines} presents a comparison of our supervised model to~\citet{jia2022cvss} in the 21-direction setting, which performs roughly similarly.
    }
    \label{table:s2st}
\end{table*}

\begin{table*}[t]
    \small
    \centering
    \begin{tabular}{c|ccccc|ccccc}
    \toprule
    wav2vec 2.0 & \multirow{2}{*}{Domain} & \multirow{2}{*}{Hours} & Multi- & Seen & Fine- & Fr & Es & Ru & Et & Lv \\
    features & & & lingual & lang. & tuning & 264h & 113h & 16h & 3h & 2h \\
    \midrule
    \midrule
    VoxPopuli & \multirow{2}{*}{out} & 21K- & \multirow{2}{*}{$\ast$} & \multirow{2}{*}{$\ast$} & none & 26.7 & 21.4 & $> \text{60}$ & $> \text{60}$ & $> \text{60}$ \\
    \citep{wang2021voxpopuli} & & 89K & & & unsup. & 21.4 & 18.3 & 25.6 & 22.4 & 27.8 \\
    \midrule
    XLS-R & \multirow{2}{*}{in+out} & \multirow{2}{*}{436K} & \multirow{2}{*}{\checkmark} & \multirow{2}{*}{\checkmark} & none & 26.1 & 21.9 & 32.8 & $> \text{60}$ & $> \text{60}$ \\
    \citep{babu2021xls} & & & & & unsup. & 23.4 & 19.0 & 28.3 & 26.4 & $> \text{60}$ \\
     \midrule
    Robust wav2vec 2.0 &\multirow{2}{*}{out} & \multirow{2}{*}{63K} & & & none & $> \text{60}$ & 29.3 & $> \text{60}$ & $> \text{60}$ & $> \text{60}$ \\
    \citep{hsu21_interspeech} & & & & & unsup. & 31.5 & 22.7 & 35.2 & 35.1 & $> \text{60}$ \\
    \bottomrule
    \end{tabular}
    \caption{Different wav2vec 2.0 features for non-English unsupervised ASR (wav2vec-U 2.0) training: validation PER on CoVoST 2 with Viterbi decoding. All models use the wav2vec 2.0 \textit{large} configuration. We unsupervisedly finetune wav2vec 2.0 models to the language and domain of interest. ``$\ast$'': Monolingual models for Fr and Es; multilingual models of similar languages for Ru, Et and Lv (trained on the Slavic, Uralic and Baltic languages in VoxPopuli, respectively).}
    \label{table:w2v2}
\end{table*}

\section{Results}

\subsection{X-En Speech-to-Text Translation}
For X-En S2TT, we consider models trained for a single language direction (bilingual) and models covering multiple directions (multilingual).
Results are reported on five translation directions into English of the CoVoST 2 benchmark and we focus on end-to-end systems but we also consider a cascade of unsupervised models.
Supervised models are purely trained on labeled data without pre-training, while as supervised models with pre-training use wav2vec and mBART models, unsupervised models also use pre-trained models but no labeled data.

Table~\ref{table:s2t_x_en} shows that unsupervised end-to-end models outperform the supervised baselines by 5.0 BLEU on average over the five translation directions of the bilingual setup. 
The supervised models represent the best supervised end-to-end models from two years ago.
These improvements are due to advances in unsupervised modeling as well as self-supervised pre-training.
The supervised models with pre-training perform generally far above the unsupervised models and shows that there is potential to improve unsupervised speech translation in the future.

The cascaded unsupervised setup performs better than the end-to-end approach for directions with little synthetic training data such as Ru-En, Et-En and Lv-En. 
This is because end-to-end models are trained on datasets comprising as little as two hours of synthetic speech translation data on which they overfit. 
Cascaded unsupervised models do not suffer under this issue because they exploit more text for unsupervised machine translation~(\autoref{table:mt}).

Supervised learning with pre-training for the bilingual setup performs better than the multilingual setup because only a single translation direction needs to be modeled and because the mBART model was pre-trained on 50 languages while as only a single language is being used in the X-En setup.


\subsection{En-X Speech-to-Text Translation}
For bilingual En-X S2TT, we compare our unsupervised models to the previous state of the art~\citep{chung2019towards} on Libri-Trans (En-Fr) and we also evaluate them on the MuST-C benchmark for En-Es and En-Ru directions.
Table~\ref{table:s2t_en_x} shows the test BLEU of our models and the baselines on both benchmarks. On Libri-Trans, our best system outperforms the previous state of the art, an alignment-based cascaded system, by 3.2 BLEU~\citep{chung2019towards}. On MuST-C, our models also achieve competitive results in this high-resource setting of around 500 hours of training data, with 3.4 BLEU and 5.5 BLEU behind the supervised baselines on En-Es and En-Ru, respectively.

\subsection{X-En Speech-to-Speech Translation}
To train a multilingual X-En speech-to-speech translation model, we combine pseudo-labeled bilingual data for multiple translation directions and use the Spec-T2 architecture, a variant of Translatotron 2. 
We build supervised Spec-T2 baselines with and without pre-training and evaluate on the CVSS-C benchmark. 
Table~\ref{table:s2st} shows that the best unsupervised system is on average only 0.8 BLEU below the supervised baseline. 
We believe that the unsupervised approach is less effective for speech-to-speech translation compared to speech-to-translation because of the increased error accumulation in the synthetic data creation process due to the addition of the unsupervised speech synthesis component to which we input unsupervised translation output which in turn is based on unsupervised speech recognition transcriptions.
Similarly to speech-to-text translation, the cascaded unsupervised model performs better than the end to end approach and this is most prominent for low-resource directions.

\begin{table}[t]
    \small
    \centering
    \begin{tabular}{l@{\hs{1.0}}c@{\hs{1.0}}c@{\hs{1.0}}c@{\hs{1.0}}c@{\hs{1.0}}c@{\hs{1.0}}c@{\hs{1.0}}c@{\hs{1.1}}c}
    \toprule
     & Fr & Es & Ru & Et & Lv & En & \multirow{2}{*}{Avg.} \\
     Duration (hrs) & 264 & 113 & 16 & 3 & 2 & 504 \\
    \midrule
    \midrule
    \multicolumn{7}{l}{\textbf{Supervised learning + pre-training}} \\
    \hspace{1.5mm}w2v2-CTC & 15.7 & \phantom{a}7.0 & \phantom{a}7.1 & 11.1 & \phantom{a}5.9 & \phantom{a}6.3 & \phantom{a}8.9 \\
    \midrule
    \multicolumn{7}{l}{\textbf{Supervised learning}} \\
    \hspace{1.5mm}Transformer$^\dagger$ & 18.3 & 16.0 & 31.4 & 65.7 & 51.8 & 12.1 & 32.6 \\
    \midrule
    \multicolumn{7}{l}{\textbf{Unsupervised learning}} \\
    \hspace{1.5mm}w2vu2-CTC & 23.2 & 10.3 & 15.7 & 17.6 & 14.8 & 12.7 & 15.7 \\
    \bottomrule
    \end{tabular}
    \caption{Speech recognition results: test WER on CoVoST 2 and MuST-C (En-Es). Semi-supervised and unsupervised models are decoded with 4-gram language model. $^\dagger$~\citet{wang-etal-2020-fairseq}.}
    \label{table:asr}
\end{table}

\begin{table}[t]
    \small
    \centering
    \begin{tabular}{l@{\hs{1.3}}c@{\hs{1.3}}c@{\hs{1.3}}c}
    \toprule
     & CVSS & Libri-Trans & MuST-C \\
    JS Divergence & 0.207 & 0.376 & 0.369 \\
    \midrule
    \midrule
    \textbf{Supervised learning} \\
    \hspace{1.5mm}Transformer & 12.8 & 15.0 & 16.8 \\
    \midrule
    \textbf{Unsupervised learning} \\
    \hspace{1.5mm}Transformer & 15.2 & 17.1 & 20.1 \\
    \bottomrule
    \end{tabular}
    \caption{Speech synthesis results: validation WER for re-synthesis on CVSS-C, Libri-Trans and MuST-C. To quantify training-inference time domain similarity, we follow~\citet{lin2022analyzing} to compute Jensen–Shannon divergence (``JSD'') on 4-gram phoneme distributions. Low JSD suggests high similarity. 
    }
    \label{table:tts}
\end{table}

\begin{table*}[t]
    \small
    \centering
    \begin{tabular}{lccccccccc}
    \toprule
     & Fr-En & Es-En & Ru-En & Et-En & Lv-En & En-Es & En-Ru & En-Fr & \multirow{3}{*}{Avg.} \\
     2.1B En text, non-En text & 428M & 379M & 849M & 46M & 68M & 379M & 849M & 428M \\
     Bitext & 207K & 79K & 12K & 1.8K & 2.3K & 259K & 259K & 47K \\
    \midrule
    \midrule
    \multicolumn{4}{l}{\textbf{Supervised learning + pre-training}} \\
    \hspace{1.5mm}mBART-FT & 46.7 & 46.0 & 48.4 & 23.3 & 29.6 & 38.7 & 23.1 & 21.5 & 34.6 \\
    \midrule
    \textbf{Supervised learning} \\
    \hspace{1.5mm}Transformer & 37.9$^\dagger$ & 36.3$^\dagger$ & 19.8$^\dagger$ & 0.3$^\dagger$ & 0.2$^\dagger$ & 33.8 & 15.8 & 17.9 & 20.3 \\
    \midrule
    \textbf{Unsupervised learning} \\
    \hspace{1.5mm}mBART-OBT & 40.1 & 43.8 & 48.6 & 19.0 & 25.0 & 38.5 & 22.2 & 22.1 & 32.4 \\
    \bottomrule
    \end{tabular}
    \caption{Machine translation results: test BLEU on CoVoST 2 (X-En), MuST-C (En-Es and En-Ru) and Libri-Trans (En-Fr). We finetune mBART model with bitext data for supervised learning and with unpaired pre-training data for unsupervised learning. $^\dagger$~\citet{wang-etal-2020-fairseq}.}
    \label{table:mt}
\end{table*}

\subsection{Speech Pre-training}
We evaluate the effectiveness of the unsupervised adaptation technique of wav2vec 2.0 models (\textsection\ref{sec:unsup_adap}) on the five non-English languages, which have less training data than English. 
We train wav2vec-U 2.0 models on CoVoST 2 with features extracted from three different wav2vec 2.0 models and their adapted versions: 1) Out-of-domain models, ``VoxPopuli''~\citep{wang2021voxpopuli}, that are trained with data in the same language (for Fr and Es) or similar languages (for Ru, Et and Lv) from the same language family subgroup; 2) a massively multilingual model for 128 languages, ``XLS-R''~\citep{babu2021xls}, whose training data contains CoVoST 2; 3) a multi-domain English model, ``robust wav2vec 2.0''~\citep{hsu21_interspeech}, where the target languages are unseen. We report validation PER on Viterbi predictions in Table~\ref{table:w2v2}. Speech pre-training on mismatched domains or languages (``VoxPopuli'' and ``robust wav2vec 2.0'') leads to training convergence failure on three low-resource languages (Ru, Et and Lv). The two languages with the least amount of data, Et and Lv, even fail with in-domain multilingual pre-training. Unsupervised adaptation significantly improves training convergence and model performance for all the 3 scenarios of speech pre-training. In an example worst case scenario, Et-En wav2vec-U 2.0 model is successfully trained with only 3 hours of Et speech data and features from an adapted out-of-language out-of-domain wav2vec 2.0 model (``robust wav2vec 2.0'').
 
\begin{table*}[t]
    \small
    \centering
    \begin{tabular}{lccccccccc}
    \toprule
     & Fr-En & Es-En & Ru-En & Et-En & Lv-En & En-Es & En-Ru & En-Fr & Avg. \\
    \midrule
    \midrule
    \multicolumn{9}{l}{\textbf{BLEU on raw text}} \\
    \hspace{1.5mm}ASR$\rightarrow$TDN$\rightarrow$MT & 24.4 & 23.4 & 27.8 & \phantom{a}8.5 & \phantom{a}7.6 & 22.0 & 10.0 & 15.4 & 17.4 \\
    \hspace{3mm}\textit{Remove TDN} & 17.2 & 18.3 & 20.7 & \phantom{a}5.7 & \phantom{a}7.8 & 17.2 & \phantom{a}8.9 & 10.4 & 13.3 \\
    \midrule
    \multicolumn{9}{l}{\textbf{BLEU on normalized text (case and punctuation removed)}} \\
    \hspace{1.5mm}ASR$\rightarrow$TDN$\rightarrow$MT & 25.0 & 23.9 & 28.7 & \phantom{a}7.9 & \phantom{a}9.5 & 23.7 & 9.4 & 15.5 & 18.0 \\
    \hspace{3mm}\textit{Remove TDN} & 23.1 & 24.1 & 26.9 & \phantom{a}7.2 & \phantom{a}9.4 & 23.1 & 9.4 & 15.1 & 17.3 \\
    \bottomrule
    \end{tabular}
    \caption{Effectiveness of text de-normalization in the unsupervised pipeline evaluated in terms of speech-to-text translation on CoVoST 2 (X-En), MuST-C (En-Es and En-Ru) and Libri-Trans (En-Fr). We report test BLEU on either raw text or normalized text. 
    TDN not only recovers case and punctuation, but also leads to better translation of content.}
    \label{table:dn}
\end{table*}

\subsection{Speech Recognition}
Next, we evaluate the performance of unsupervised speech recognition in our setting.
We decode our pre-trained supervised baselines (``w2v2-CTC'') and unsupervised models (``w2vu2-CTC'') with 4-gram language model. They are compared with previous un-pre-trained supervised baselines~\citep{wang-etal-2020-fairseq} on CoVoST 2 and MuST-C (for En), whose results (test WER) can be found in Table~\ref{table:asr}. We see that our unsupervised end-to-end models outperform un-pre-trained supervised baselines on all six languages with an average 16.9 WER reduction over the supervised one. 
Unsupervised ASR works best for languages with little labeled data due to the use of pre-trained features and advances in unsupervised algorithms.

\subsection{Speech Synthesis}
\label{sec:results_tts}
In our unsupervised setting, the target speech data does not share the same domain as the source one. This realistic setting leads to training-inference time domain mismatch on TTS models. We evaluate the effects of this mismatch by a re-synthesis task on 3 different datasets: CVSS-C (from It-En), Libri-Trans and MuST-C. We synthesize speech using validation texts and report WER on the ASR transcription of the synthesized speech. To quantize domain similarity, we follow~\citet{lin2022analyzing} to compute Jensen–Shannon divergence (``JSD'') on 4-gram phoneme distributions, where low JSD suggests high similarity. Table~\ref{table:tts} shows the results.
We see that both supervised and unsupervised models have higher WER on less similar domains (Libri-Trans and MuST-C).

\subsection{Machine Translation}
We evaluate our unsupervised models (``mBART-OBT'') on the CoVoST 2, MuST-C and Libri-Trans benchmarks with test BLEU. For comparison, we also build supervised Transformer baselines (``Transformer'') and supervised mBART baselines (``mBART-FT''). Results are shown in Table~\ref{table:mt}. We observe that our unsupervised models outperform supervised baselines by 12.1 BLEU on average over the eight considered translation directions. They are behind supervised baselines by only 2.2 BLEU on average. In contrast to supervised baselines that leverage in-domain paired data, the unsupervised models use unpaired CC100 data which is web data.

\subsection{Text De-normalization}
We verify the effectiveness of text de-normalization (TDN) by ablating it in the unsupervised cascaded pipeline. In
Table~\ref{table:dn}, we show test BLEU calculated on either raw text ($\text{BLEU}_\text{raw}$) or normalized text ($\text{BLEU}_\text{norm}$) for the ablation. We see that TDN improves $\text{BLEU}_\text{raw}$ greatly by 4.1 on average over all the directions. From the improvements on $\text{BLEU}_\text{norm}$, we conclude that TDN not only recovers case and punctuation, but also improves translation of the content.


\section{Conclusion}

In this paper, we present a simple and effective approach to unsupervised speech-to-text translation (S2TT) and speech-to-speech translation (S2ST). 
Our S2TT systems outperform the previous state of the art on Libri-Trans by 3.2 BLEU as well as the best supervised end-to-end models (without pre-training) on CoVoST 2 from only two years ago by an average of 5.0 BLEU over five translation directions into English. 
Our S2TT and S2ST systems also perform competitively on the MuST-C and CVSS-C benchmarks.

\section*{Acknowledgments}
We thank Alexei Baevski, Andy Chung, Alexis Conneau, Hongyu Gong, Jiatao Gu and Sravya Popuri for helpful discussions.

\bibliography{custom}

\begin{thebibliography}{48}
\expandafter\ifx\csname natexlab\endcsname\relax\def\natexlab#1{#1}\fi

\bibitem[{Artetxe et~al.(2018)Artetxe, Labaka, Agirre, and
  Cho}]{artetxe2018unsupervised}
Mikel Artetxe, Gorka Labaka, Eneko Agirre, and Kyunghyun Cho. 2018.
\newblock Unsupervised neural machine translation.
\newblock In \emph{International Conference on Learning Representations}.

\bibitem[{Babu et~al.(2021)Babu, Wang, Tjandra, Lakhotia, Xu, Goyal, Singh, von
  Platen, Saraf, Pino et~al.}]{babu2021xls}
Arun Babu, Changhan Wang, Andros Tjandra, Kushal Lakhotia, Qiantong Xu, Naman
  Goyal, Kritika Singh, Patrick von Platen, Yatharth Saraf, Juan Pino, et~al.
  2021.
\newblock Xls-r: Self-supervised cross-lingual speech representation learning
  at scale.
\newblock \emph{arXiv preprint arXiv:2111.09296}.

\bibitem[{Baevski et~al.(2021)Baevski, Hsu, Conneau, and
  Auli}]{baevski2021unsupervised}
Alexei Baevski, Wei-Ning Hsu, Alexis Conneau, and Michael Auli. 2021.
\newblock Unsupervised speech recognition.
\newblock \emph{Advances in Neural Information Processing Systems},
  34:27826--27839.

\bibitem[{Baevski et~al.(2020)Baevski, Zhou, Mohamed, and
  Auli}]{baevski2020wav}
Alexei Baevski, Yuhao Zhou, Abdelrahman Mohamed, and Michael Auli. 2020.
\newblock wav2vec 2.0: {A} framework for self-supervised learning of speech
  representations.
\newblock In \emph{Proc. of NeurIPS}.

\bibitem[{Bansal et~al.(2018)Bansal, Kamper, Livescu, Lopez, and
  Goldwater}]{bansal2018low}
Sameer Bansal, Herman Kamper, Karen Livescu, Adam Lopez, and Sharon Goldwater.
  2018.
\newblock Low-resource speech-to-text translation.
\newblock \emph{Proc. Interspeech 2018}, pages 1298--1302.

\bibitem[{Bansal et~al.(2019)Bansal, Kamper, Livescu, Lopez, and
  Goldwater}]{bansal2019pre}
Sameer Bansal, Herman Kamper, Karen Livescu, Adam Lopez, and Sharon Goldwater.
  2019.
\newblock Pre-training on high-resource speech recognition improves
  low-resource speech-to-text translation.
\newblock In \emph{Proceedings of the 2019 Conference of the North American
  Chapter of the Association for Computational Linguistics: Human Language
  Technologies, Volume 1 (Long and Short Papers)}, pages 58--68.

\bibitem[{Bansal et~al.(2017)Bansal, Kamper, Lopez, and
  Goldwater}]{bansal2017towards}
Sameer Bansal, Herman Kamper, Adam Lopez, and Sharon Goldwater. 2017.
\newblock Towards speech-to-text translation without speech recognition.
\newblock In \emph{Proceedings of the 15th Conference of the European Chapter
  of the Association for Computational Linguistics: Volume 2, Short Papers},
  pages 474--479.

\bibitem[{Bentivogli et~al.(2021)Bentivogli, Cettolo, Gaido, Karakanta,
  Martinelli, Negri, and Turchi}]{bentivogli2021cascade}
Luisa Bentivogli, Mauro Cettolo, Marco Gaido, Alina Karakanta, Alberto
  Martinelli, Matteo Negri, and Marco Turchi. 2021.
\newblock Cascade versus direct speech translation: Do the differences still
  make a difference?
\newblock In \emph{Proceedings of the 59th Annual Meeting of the Association
  for Computational Linguistics and the 11th International Joint Conference on
  Natural Language Processing (Volume 1: Long Papers)}, pages 2873--2887.

\bibitem[{B{\'e}rard et~al.(2016)B{\'e}rard, Pietquin, Besacier, and
  Servan}]{berard2016listen}
Alexandre B{\'e}rard, Olivier Pietquin, Laurent Besacier, and Christophe
  Servan. 2016.
\newblock Listen and translate: A proof of concept for end-to-end
  speech-to-text translation.
\newblock In \emph{NIPS Workshop on end-to-end learning for speech and audio
  processing}.

\bibitem[{Bernard(2015)}]{phonemizer2015}
Mathieu Bernard. 2015.
\newblock Phonemizer.
\newblock https://github.com/bootphon/phonemizer.

\bibitem[{Chen et~al.(2019)Chen, Tsai, Liu, Lee, and shan
  Lee}]{chen2019completely}
Kuan-Yu Chen, Che-Ping Tsai, Da-Rong Liu, Hung-Yi Lee, and Lin shan Lee. 2019.
\newblock Completely unsupervised speech recognition by a generative
  adversarial network harmonized with iteratively refined hidden markov models.
\newblock In \emph{Proc. of Interspeech}.

\bibitem[{Cheng et~al.(2021)Cheng, Lee, and Wang}]{cheng21_interspeech}
Yao-Fei Cheng, Hung-Shin Lee, and Hsin-Min Wang. 2021.
\newblock \href {https://doi.org/10.21437/Interspeech.2021-526} {{AlloST:
  Low-Resource Speech Translation Without Source Transcription}}.
\newblock In \emph{Proc. Interspeech 2021}, pages 2252--2256.

\bibitem[{Chung et~al.(2019)Chung, Weng, Tong, and Glass}]{chung2019towards}
Yu-An Chung, Wei-Hung Weng, Schrasing Tong, and James Glass. 2019.
\newblock Towards unsupervised speech-to-text translation.
\newblock In \emph{ICASSP 2019-2019 IEEE International Conference on Acoustics,
  Speech and Signal Processing (ICASSP)}, pages 7170--7174. IEEE.

\bibitem[{Conneau et~al.(2020)Conneau, Khandelwal, Goyal, Chaudhary, Wenzek,
  Guzm{\'a}n, Grave, Ott, Zettlemoyer, and
  Stoyanov}]{conneau-etal-2020-unsupervised}
Alexis Conneau, Kartikay Khandelwal, Naman Goyal, Vishrav Chaudhary, Guillaume
  Wenzek, Francisco Guzm{\'a}n, Edouard Grave, Myle Ott, Luke Zettlemoyer, and
  Veselin Stoyanov. 2020.
\newblock Unsupervised cross-lingual representation learning at scale.
\newblock In \emph{Proceedings of the 58th Annual Meeting of the Association
  for Computational Linguistics}, pages 8440--8451, Online. Association for
  Computational Linguistics.

\bibitem[{Conneau et~al.(2018)Conneau, Lample, Ranzato, Denoyer, and
  J{\'{e}}gou}]{conneau2018unsupmt}
Alexis Conneau, Guillaume Lample, Marc'Aurelio Ranzato, Ludovic Denoyer, and
  Herv{\'{e}} J{\'{e}}gou. 2018.
\newblock Word translation without parallel data.
\newblock \emph{Proc. of ICLR}.

\bibitem[{Di~Gangi et~al.(2019{\natexlab{a}})Di~Gangi, Cattoni, Bentivogli,
  Negri, and Turchi}]{di2019must}
Mattia~A Di~Gangi, Roldano Cattoni, Luisa Bentivogli, Matteo Negri, and Marco
  Turchi. 2019{\natexlab{a}}.
\newblock Must-c: a multilingual speech translation corpus.
\newblock In \emph{2019 Conference of the North American Chapter of the
  Association for Computational Linguistics: Human Language Technologies},
  pages 2012--2017. Association for Computational Linguistics.

\bibitem[{Di~Gangi et~al.(2019{\natexlab{b}})Di~Gangi, Negri, and
  Turchi}]{di2019adapting}
Mattia~A Di~Gangi, Matteo Negri, and Marco Turchi. 2019{\natexlab{b}}.
\newblock Adapting transformer to end-to-end spoken language translation.
\newblock In \emph{INTERSPEECH 2019}, pages 1133--1137. International Speech
  Communication Association (ISCA).

\bibitem[{Duong et~al.(2016)Duong, Anastasopoulos, Chiang, Bird, and
  Cohn}]{duong2016attentional}
Long Duong, Antonios Anastasopoulos, David Chiang, Steven Bird, and Trevor
  Cohn. 2016.
\newblock An attentional model for speech translation without transcription.
\newblock In \emph{Proceedings of the 2016 Conference of the North American
  Chapter of the Association for Computational Linguistics: Human Language
  Technologies}, pages 949--959.

\bibitem[{Graves et~al.(2006)Graves, Fern{\'a}ndez, Gomez, and
  Schmidhuber}]{graves2006connectionist}
Alex Graves, Santiago Fern{\'a}ndez, Faustino Gomez, and J{\"u}rgen
  Schmidhuber. 2006.
\newblock Connectionist temporal classification: labelling unsegmented sequence
  data with recurrent neural networks.
\newblock In \emph{Proceedings of the 23rd international conference on Machine
  learning}, pages 369--376.

\bibitem[{Hsu et~al.(2021)Hsu, Sriram, Baevski, Likhomanenko, Xu, Pratap, Kahn,
  Lee, Collobert, Synnaeve, and Auli}]{hsu21_interspeech}
Wei-Ning Hsu, Anuroop Sriram, Alexei Baevski, Tatiana Likhomanenko, Qiantong
  Xu, Vineel Pratap, Jacob Kahn, Ann Lee, Ronan Collobert, Gabriel Synnaeve,
  and Michael Auli. 2021.
\newblock {Robust wav2vec 2.0: Analyzing Domain Shift in Self-Supervised
  Pre-Training}.
\newblock In \emph{Proc. Interspeech 2021}, pages 721--725.

\bibitem[{Jia et~al.(2022{\natexlab{a}})Jia, Ramanovich, Remez, and
  Pomerantz}]{pmlr-v162-jia22b}
Ye~Jia, Michelle~Tadmor Ramanovich, Tal Remez, and Roi Pomerantz.
  2022{\natexlab{a}}.
\newblock Translatotron 2: High-quality direct speech-to-speech translation
  with voice preservation.
\newblock In \emph{Proceedings of the 39th International Conference on Machine
  Learning}, volume 162 of \emph{Proceedings of Machine Learning Research},
  pages 10120--10134. PMLR.

\bibitem[{Jia et~al.(2022{\natexlab{b}})Jia, Ramanovich, Wang, and
  Zen}]{jia2022cvss}
Ye~Jia, Michelle~Tadmor Ramanovich, Quan Wang, and Heiga Zen.
  2022{\natexlab{b}}.
\newblock Cvss corpus and massively multilingual speech-to-speech translation.
\newblock \emph{arXiv preprint arXiv:2201.03713}.

\bibitem[{Jia et~al.(2019)Jia, Weiss, Biadsy, Macherey, Johnson, Chen, and
  Wu}]{Jia2019DirectST}
Ye~Jia, Ron~J. Weiss, Fadi Biadsy, Wolfgang Macherey, Melvin Johnson, Z.~Chen,
  and Yonghui Wu. 2019.
\newblock Direct speech-to-speech translation with a sequence-to-sequence
  model.
\newblock In \emph{INTERSPEECH}.

\bibitem[{Kano et~al.(2021)Kano, Sakti, and Nakamura}]{kano2021transformer}
Takatomo Kano, Sakriani Sakti, and Satoshi Nakamura. 2021.
\newblock Transformer-based direct speech-to-speech translation with
  transcoder.
\newblock In \emph{2021 IEEE Spoken Language Technology Workshop (SLT)}, pages
  958--965. IEEE.

\bibitem[{Kocabiyikoglu et~al.(2018)Kocabiyikoglu, Besacier, and
  Kraif}]{kocabiyikoglu2018augmenting}
Ali~Can Kocabiyikoglu, Laurent Besacier, and Olivier Kraif. 2018.
\newblock Augmenting librispeech with french translations: A multimodal corpus
  for direct speech translation evaluation.
\newblock In \emph{Proceedings of the Eleventh International Conference on
  Language Resources and Evaluation (LREC 2018)}.

\bibitem[{Lample et~al.(2018)Lample, Conneau, Denoyer, and
  Ranzato}]{lample2018unsupervised}
Guillaume Lample, Alexis Conneau, Ludovic Denoyer, and Marc'Aurelio Ranzato.
  2018.
\newblock Unsupervised machine translation using monolingual corpora only.
\newblock In \emph{International Conference on Learning Representations}.

\bibitem[{Lewis et~al.(2022)Lewis, Simon, and Fennig}]{lewis2022ethnologue}
M.~Paul Lewis, Gary~F. Simon, and Charles~D. Fennig. 2022.
\newblock Ethnologue: Languages of the world, 25th edition.
\newblock Online version: \url{http://www.ethnologue.com}.

\bibitem[{Li et~al.(2019)Li, Liu, Liu, Zhao, and
  Liu}]{Li_Liu_Liu_Zhao_Liu_2019}
Naihan Li, Shujie Liu, Yanqing Liu, Sheng Zhao, and Ming Liu. 2019.
\newblock Neural speech synthesis with transformer network.
\newblock \emph{Proceedings of the AAAI Conference on Artificial Intelligence},
  33(01):6706--6713.

\bibitem[{Li et~al.(2021)Li, Wang, Tang, Tran, Tang, Pino, Baevski, Conneau,
  and Auli}]{li-etal-2021-multilingual}
Xian Li, Changhan Wang, Yun Tang, Chau Tran, Yuqing Tang, Juan Pino, Alexei
  Baevski, Alexis Conneau, and Michael Auli. 2021.
\newblock Multilingual speech translation from efficient finetuning of
  pretrained models.
\newblock In \emph{Proceedings of the 59th Annual Meeting of the Association
  for Computational Linguistics and the 11th International Joint Conference on
  Natural Language Processing (Volume 1: Long Papers)}, pages 827--838, Online.
  Association for Computational Linguistics.

\bibitem[{Lin et~al.(2022)Lin, Hsu, Liu, Lee, and Tsao}]{lin2022analyzing}
Guan-Ting Lin, Chan-Jan Hsu, Da-Rong Liu, Hung-Yi Lee, and Yu~Tsao. 2022.
\newblock Analyzing the robustness of unsupervised speech recognition.
\newblock In \emph{ICASSP 2022-2022 IEEE International Conference on Acoustics,
  Speech and Signal Processing (ICASSP)}, pages 8202--8206. IEEE.

\bibitem[{Liu et~al.(2022{\natexlab{a}})Liu, Hsu, Auli, and
  Baevski}]{liu2022towards}
Alexander~H Liu, Wei-Ning Hsu, Michael Auli, and Alexei Baevski.
  2022{\natexlab{a}}.
\newblock Towards end-to-end unsupervised speech recognition.
\newblock \emph{arXiv preprint arXiv:2204.02492}.

\bibitem[{Liu et~al.(2022{\natexlab{b}})Liu, Lai, Hsu, Auli, Baevskiv, and
  Glass}]{liu2022simple}
Alexander~H Liu, Cheng-I~Jeff Lai, Wei-Ning Hsu, Michael Auli, Alexei Baevskiv,
  and James Glass. 2022{\natexlab{b}}.
\newblock Simple and effective unsupervised speech synthesis.
\newblock \emph{arXiv preprint arXiv:2204.02524}.

\bibitem[{Liu et~al.(2018)Liu, Chen, Lee, and shan Lee}]{liu2018completely}
Da-Rong Liu, Kuan-Yu Chen, Hung-Yi Lee, and Lin shan Lee. 2018.
\newblock Completely unsupervised phoneme recognition by adversarially learning
  mapping relationships from audio embeddings.
\newblock \emph{Proc. of Interspeech}.

\bibitem[{Liu et~al.(2020)Liu, Gu, Goyal, Li, Edunov, Ghazvininejad, Lewis, and
  Zettlemoyer}]{liu2020multilingual}
Yinhan Liu, Jiatao Gu, Naman Goyal, Xian Li, Sergey Edunov, Marjan
  Ghazvininejad, Mike Lewis, and Luke Zettlemoyer. 2020.
\newblock Multilingual denoising pre-training for neural machine translation.
\newblock \emph{Transactions of the Association for Computational Linguistics},
  8:726--742.

\bibitem[{Mohri(1997)}]{mohri1997finite}
Mehryar Mohri. 1997.
\newblock Finite-state transducers in language and speech processing.
\newblock \emph{Computational linguistics}, 23(2):269--311.

\bibitem[{Ni et~al.(2022)Ni, Wang, Gao, Qian, Zhang, Chang, and
  Hasegawa-Johnson}]{ni2022unsupervised}
Junrui Ni, Liming Wang, Heting Gao, Kaizhi Qian, Yang Zhang, Shiyu Chang, and
  Mark Hasegawa-Johnson. 2022.
\newblock Unsupervised text-to-speech synthesis by unsupervised automatic
  speech recognition.
\newblock \emph{arXiv preprint arXiv:2203.15796}.

\bibitem[{Park(2019)}]{g2pE2019}
Jongseok Park, Kyubyong \&~Kim. 2019.
\newblock g2pe.
\newblock https://github.com/Kyubyong/g2p.

\bibitem[{Ren et~al.(2020)Ren, Liu, Tan, Zhang, Qin, Zhao, and
  Liu}]{ren2020simulspeech}
Yi~Ren, Jinglin Liu, Xu~Tan, Chen Zhang, Tao Qin, Zhou Zhao, and Tie-Yan Liu.
  2020.
\newblock Simulspeech: End-to-end simultaneous speech to text translation.
\newblock In \emph{Proceedings of the 58th Annual Meeting of the Association
  for Computational Linguistics}, pages 3787--3796.

\bibitem[{Schneider et~al.(2019)Schneider, Baevski, Collobert, and
  Auli}]{schneider2019wav2vec}
Setffen Schneider, Alexei Baevski, Ronan Collobert, and Michael Auli. 2019.
\newblock wav2vec: Unsupervised pre-training for speech recognition.
\newblock In \emph{Proc. of Interspeech}.

\bibitem[{Vaswani et~al.(2017)Vaswani, Shazeer, Parmar, Uszkoreit, Jones,
  Gomez, Kaiser, and Polosukhin}]{vaswani2017attention}
Ashish Vaswani, Noam Shazeer, Niki Parmar, Jakob Uszkoreit, Llion Jones,
  Aidan~N Gomez, {\L}ukasz Kaiser, and Illia Polosukhin. 2017.
\newblock Attention is all you need.
\newblock \emph{Advances in neural information processing systems}, 30.

\bibitem[{Vila et~al.(2018)Vila, Escolano, Fonollosa, and
  Costa-Jussa}]{vila2018end}
Laura~Cross Vila, Carlos Escolano, Jos{\'e}~AR Fonollosa, and Marta~R
  Costa-Jussa. 2018.
\newblock End-to-end speech translation with the transformer.
\newblock In \emph{IberSPEECH}, pages 60--63.

\bibitem[{Vincent et~al.(2010)Vincent, Larochelle, Lajoie, Bengio, Manzagol,
  and Bottou}]{vincent2010stacked}
Pascal Vincent, Hugo Larochelle, Isabelle Lajoie, Yoshua Bengio, Pierre-Antoine
  Manzagol, and L{\'e}on Bottou. 2010.
\newblock Stacked denoising autoencoders: Learning useful representations in a
  deep network with a local denoising criterion.
\newblock \emph{Journal of machine learning research}, 11(12).

\bibitem[{Wang et~al.(2021{\natexlab{a}})Wang, Riviere, Lee, Wu, Talnikar,
  Haziza, Williamson, Pino, and Dupoux}]{wang2021voxpopuli}
Changhan Wang, Morgane Riviere, Ann Lee, Anne Wu, Chaitanya Talnikar, Daniel
  Haziza, Mary Williamson, Juan Pino, and Emmanuel Dupoux. 2021{\natexlab{a}}.
\newblock Voxpopuli: A large-scale multilingual speech corpus for
  representation learning, semi-supervised learning and interpretation.
\newblock In \emph{Proceedings of the 59th Annual Meeting of the Association
  for Computational Linguistics and the 11th International Joint Conference on
  Natural Language Processing (Volume 1: Long Papers)}, pages 993--1003.

\bibitem[{Wang et~al.(2020)Wang, Tang, Ma, Wu, Okhonko, and
  Pino}]{wang-etal-2020-fairseq}
Changhan Wang, Yun Tang, Xutai Ma, Anne Wu, Dmytro Okhonko, and Juan Pino.
  2020.
\newblock Fairseq {S}2{T}: Fast speech-to-text modeling with fairseq.
\newblock In \emph{Proceedings of the 1st Conference of the Asia-Pacific
  Chapter of the Association for Computational Linguistics and the 10th
  International Joint Conference on Natural Language Processing: System
  Demonstrations}, pages 33--39, Suzhou, China. Association for Computational
  Linguistics.

\bibitem[{Wang et~al.(2021{\natexlab{b}})Wang, Wu, Gu, and
  Pino}]{wang2021covost}
Changhan Wang, Anne Wu, Jiatao Gu, and Juan Pino. 2021{\natexlab{b}}.
\newblock Covost 2 and massively multilingual speech translation.
\newblock In \emph{Interspeech}, pages 2247--2251.

\bibitem[{Weiss et~al.(2017)Weiss, Chorowski, Jaitly, Wu, and
  Chen}]{weiss2017sequence}
Ron~J Weiss, Jan Chorowski, Navdeep Jaitly, Yonghui Wu, and Zhifeng Chen. 2017.
\newblock Sequence-to-sequence models can directly translate foreign speech.
\newblock \emph{Proc. Interspeech 2017}, pages 2625--2629.

\bibitem[{Wu et~al.(2021)Wu, Li, Wang, Meng, Qin, Chen, Zhang, Liu
  et~al.}]{wu2021r}
Lijun Wu, Juntao Li, Yue Wang, Qi~Meng, Tao Qin, Wei Chen, Min Zhang, Tie-Yan
  Liu, et~al. 2021.
\newblock R-drop: Regularized dropout for neural networks.
\newblock \emph{Advances in Neural Information Processing Systems},
  34:10890--10905.

\bibitem[{Yeh et~al.(2019)Yeh, Chen, Yu, and Yu}]{yeh2018unsupervised}
Chih-Kuan Yeh, Jianshu Chen, Chengzhu Yu, and Dong Yu. 2019.
\newblock Unsupervised speech recognition via segmental empirical output
  distribution matching.
\newblock In \emph{Proc. of ICLR}.

\end{thebibliography}

\clearpage
\appendix

\section{Appendix}
\subsection{Comparison of our CVSS-C supervised baseline to previous work}
\label{app:cvss_baselines}
\begin{table}[h]
    \small
    \centering
    \begin{tabular}{l@{\hs{1.0}}c@{\hs{1.3}}c@{\hs{1.3}}c@{\hs{1.3}}c@{\hs{1.3}}c@{\hs{1.3}}c}
    \toprule
    X-En direction & Fr & Es & Ru & Et & Lv & \multirow{1}{*}{Avg.} \\
    \midrule
    \midrule
    \multicolumn{7}{l}{\textbf{Evaluated by a proprietary ASR}} \\
    \hspace{1.5mm}\citet{jia2022cvss} & 32.4 & 33.4 & 23.2 & \phantom{a}3.2 & \phantom{a}2.8 & 19.0 \\
    \midrule
    \multicolumn{7}{l}{\textbf{Evaluated by an open-source ASR}} \\
    \hspace{1.5mm}Ours & 33.8 & 34.6 & 29.4 & \phantom{a}3.1 & \phantom{a}3.2 & 20.8 \\ 
    \bottomrule
    \end{tabular}
    \caption{Multilingual supervised baselines on CVSS-C for translating 21 languages into English. We report test BLEU on ASR transcription of the translated speech.
    }
    \label{table:s2st_sup}
\end{table}

For evaluation of CVSS-C models, we use an open-source English ASR model\footnote{https://github.com/facebookresearch/fairseq/tree/main/\newline
examples/wav2vec (``Wav2Vec 2.0 Large (LV-60) + Self Training'')} to transcribe translated speech for BLEU calculation. The previous work~\citep{jia2022cvss}, however, used transcription from a proprietary ASR model which we do not have access to. As a result, BLEU numbers reported for our model and the previous work are not directly comparable, but the small difference suggests that the two models perform roughly similarly.

\subsection{Data Overview for Supervised Learning and Unsupervised Learning}
\label{app:data_overview}
\begin{table}[h]
    \small
    \centering
    \begin{tabular}{l@{\hs{1.0}}r@{\hs{1.0}}r@{\hs{1.0}}r@{\hs{1.0}}r@{\hs{1.0}}r}
    \toprule
     & Fr-En & Es-En & Ru-En & Et-En & Lv-En \\
    \midrule
    \midrule
    \multicolumn{5}{l}{\textbf{Supervised learning}} \\
    Src. paired speech & 264 & 113 & 16 & 3 & 2 \\
    Src. paired text & 207K & 79K & 12K & 1.8K & 2.3K \\
    Tgt. paired speech & 174 & 70 & 13 & 3 & 1 \\
    Tgt. paired text & 207K & 79K & 12K & 1.8K & 2.3K \\
    \multicolumn{5}{l}{\textbf{Unsupervised learning}} \\
    Src. speech & 23K & 21K & 89K & 43K & 28K \\
    Src. text & 428M & 379M & 849M & 46M & 68M \\
    Tgt. speech & 29 & 29 & 29 & 29 & 29 \\
    Tgt. text & 2.1B & 2.1B & 2.1B & 2.1B & 2.1B \\
    \bottomrule
    \toprule
    & En-Es & En-Ru & En-Fr \\
    \midrule
    \midrule
    \multicolumn{5}{l}{\textbf{Supervised learning}} \\
    Src. paired speech & 504 & 489 & 100 \\
    Src. paired text & 259K & 259K & 47K \\
    Tgt. paired text & 259K & 259K & 47K \\
    \multicolumn{5}{l}{\textbf{Unsupervised learning}} \\
    Src. speech & 63K & 63K & 63K \\
    Src. text & 2.1B & 2.1B & 2.1B \\
    Tgt. text & 379M & 849M & 428M \\
    \bottomrule
    \end{tabular}
    \caption{Overview of the speech data (hours) and text data (sentences) used in supervised learning and unsupervised learning.}
    \label{table:data}
\end{table}

Table~\ref{table:data} provides an overview for the speech and text data used in supervised learning and unsupervised learning.

\end{document}